\documentclass[runningheads]{llncs}
\usepackage{graphicx}
\usepackage{comment}
\usepackage{amsmath,amssymb} 
\usepackage{color}
\usepackage{subcaption}
\usepackage{multirow}
\usepackage{caption}
\usepackage{float}

\newcommand*{\affmark}[1][*]{\textsuperscript{#1}}

\begin{document}
\pagestyle{headings}
\mainmatter
\def\ECCVSubNumber{4031}  

\title{2D Wasserstein Loss for Robust Facial Landmark Detection} 

\titlerunning{2D Wasserstein Loss for Robust FLD}
%
\author{
Yongzhe Yan\affmark[1] \ \ \ \ Stefan Duffner\affmark[2]\ \ \ \ Priyanka Phutane\affmark[1]\ \ \ \ Anthony Berthelier\affmark[1]\ \ \ \ Christophe Blanc\affmark[1]\ \ \ \
Christophe Garcia\affmark[2]\ \ \ \ Thierry Chateau\affmark[1]
}

\authorrunning{Y. Yan et al.}
%
\institute{Universit\'e Clermont Auvergne, CNRS, SIGMA, Institut Pascal, Clermont-Ferrand, France \and
Universit\'e de Lyon, CNRS, INSA-Lyon, LIRIS, UMR5205, France 
\email{yongzhe.yan@etu.uca.fr}}
\maketitle

\begin{abstract}
The recent performance of facial landmark detection has been significantly improved by using deep Convolutional Neural Networks (CNNs), especially the Heatmap Regression Models (HRMs). 
Although their performance on common benchmark datasets has reached a high level, the robustness of these models still remains a challenging problem in the practical use under noisy conditions of realistic environments.
Contrary to most existing work focusing on the design of new models, we argue that improving the robustness requires rethinking many other aspects, including the use of datasets, the format of landmark annotation, the evaluation metric as well as the training and detection algorithm itself.
In this paper, we propose a novel method for robust facial landmark detection, using a loss function based on the 2D Wasserstein distance combined with a new landmark coordinate sampling relying on the barycenter of the individual probability distributions.
Our method can be plugged-and-play on most state-of-the-art HRMs with neither additional complexity nor structural modifications of the models. 
Further, with the large performance increase,
we found that current evaluation metrics can no longer fully reflect the robustness of these models. 
Therefore, we propose several improvements to the standard evaluation protocol.
Extensive experimental results on both traditional evaluation metrics and our evaluation metrics demonstrate that our approach significantly improves the robustness of state-of-the-art facial landmark detection models.
\end{abstract}

\section{Introduction}
Facial landmark detection has been a highly active research topic in the last decade and
plays an important role in most face image analysis applications e.g. face recognition, face editing and face 3D reconstructions, etc.. 
Recently, neural network-based Heatmap Regression Models (HRMs) outperform other methods due to their strong capability of handling large pose variations. 
Unlike Coordinate Regression CNNs which directly estimate the numerical coordinates using fully-connected layers at the output, HRMs usually adopt a fully-convolutional CNN structure. 
The training targets of HRMs are heatmaps composed of Gaussian distributions centered at the ground truth position of each landmark~\cite{duffner_ispa05}.
Recently, HRMs have brought the performance on current benchmarks to a very high level. However, maintaining robustness is still challenging in the practical use, especially with video streams that involve motion blur, self-occlusions, changing lighting conditions, etc. 
\par

We think that the use of geometric information is the key to further improve the robustness.
As faces are 3D objects bound to some physical constraints, there exists a natural correlation between landmark positions in the 2D images.
This correlation contains important but implicit geometric information. 
However, the $L2$ loss that is comonly used to train state-of-the-art HRMs is not able to exploit this geometric information. Hence, we propose a new loss function based on the 2D Wasserstein distance (loss).
\par

\begin{figure}[t]
\centering
\begin{center}
\includegraphics[width=0.6\textwidth]{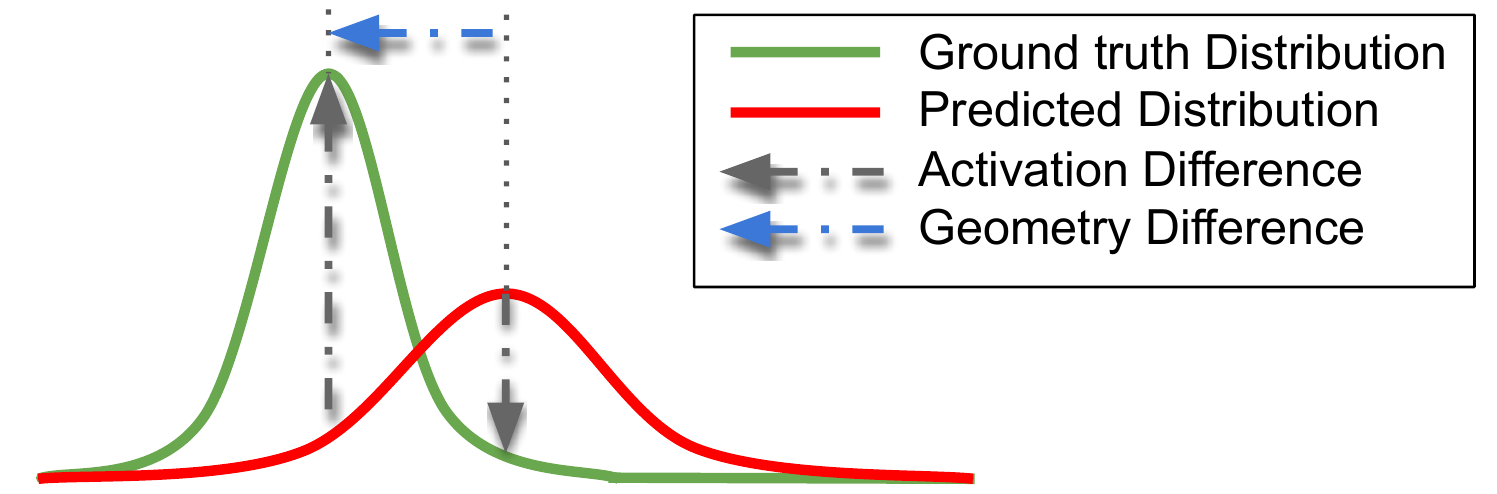}
\end{center}
   \caption{An illustration of the Wasserstein loss between two 1D distributions. Standard $L2$ loss only considers the ``activation'' difference (point-wise value difference, vertical gray arrows), whereas the Wasserstein loss takes into account both the activation and the geometry differences (distance between points, horizontal blue arrow).}
\label{fig:wloss1D}
\end{figure}

The Wasserstein distance, a.k.a. Earth Mover's Distance, is a widely used metric in Optimal Transport Theory~\cite{villani2008optimal}. 
It measures the distance between two probability distributions and has an intuitive interpretation. 
If we consider each probability distribution as a pile of earth, this distance represents the minimum effort to move the earth from one pile to the other.  
Unlike other measurements such as $L2$, Kullback-Leibler divergence and Jensen-Shannon divergence, the most appealing property of the Wasserstein distance is its sensitivity to the geometry (see Fig.~\ref{fig:wloss1D}).

The contribution of this article is two-fold:
\begin{itemize}
\item[$\bullet$] We propose a novel method based on the Wasserstein loss to significantly improve the robustness of facial landmark detection.\par
\item[$\bullet$] We propose several modifications to the current evaluation metrics to reflect the robustness of the state-of-the-art methods more effectively.\par
\end{itemize}

\section{Context \& Motivation}
\label{sec:emp_study}
\textbf{Related work: }
Robust facial landmark detection in images is a long-standing research topic. Numerous works~\cite{burgos2013robust,asthana2013robust,smith2014nonparametric,zhao2013cascaded,yu2014consensus,wu2017simultaneous,wu2017simultaneous,zhou2013exemplar,feng2017face,yang2015robust,wu2015robust,baltrusaitis2013constrained} propose methods to improve the overall detection robustness on Active
Appearance Models~\cite{cootes2001active}, Constrained Local Models (CLM)~\cite{cristinacce2006feature}, Exemplars-based Models~\cite{belhumeur2013localizing} and Cascaded Regression Models~\cite{dollar2010cascaded}. 
Specifically, the heatmap used in HRMs is conceptually connected to the response map used in CLMs in terms of local activation.
Both RLMS~\cite{saragih2011deformable} and DRMF~\cite{asthana2013robust} made effort to alleviate the robustness problem in CLM models.

These approaches have been superseded more recently with the advent of very powerful deep neural network models. 
In this context, several works have been proposed for robust facial landmark detection~\cite{zhu2019robust,xiao2016robust,kowalski2017deep,merget2018robust,yang2017stacked,feng2018wing,wang2019adaptive} by carefully designing CNN models, by balancing the data distribution and other specific techniques.
Chen et al.~\cite{Chen2019NIPS_structured} combined Conditional Random Field with the CNNs to produce structured probabilistic prediction.
FAB~\cite{Sun_2019_FAB} introduced Structure-aware Deblurring to enhance the robustness against motion blur.

\begin{figure}[t]
\centering
\begin{center}
    \includegraphics[width=\textwidth]{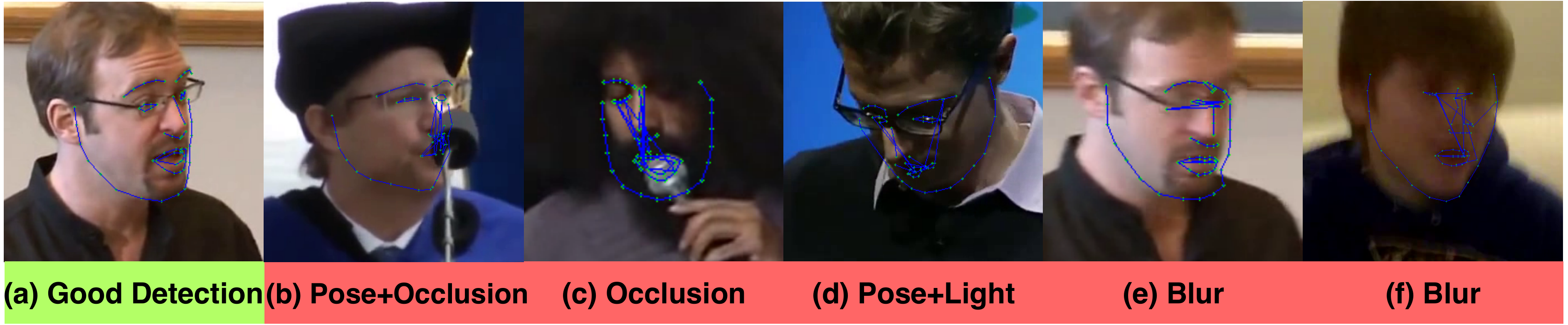}
\end{center}
\caption{Examples of HRNet detection on 300VW-S3.}
\label{fig:robustness problem}
\end{figure}

\textbf{Robustness problem of HRMs: } Figure~\ref{fig:robustness problem} shows some example results of the state-of-the-art method HRNet~\cite{sun2019high}.
HRNet can handle most of the challenging situations (e.g.\ Fig.~\ref{fig:robustness problem} (a)). 
However, we observed that a well-trained HRNet still has difficulties in the practical use when facing extreme poses (Fig.~\ref{fig:robustness problem} (b)(d)), heavy occlusions (Fig.~\ref{fig:robustness problem} (b)(c)) and motion blur (Fig.~\ref{fig:robustness problem} (e)(f)). \par

These observed robustness issues are rather specific to HRMs.
When using Cascaded Regression Models or Coordinate Regression CNNs, even if the prediction is poor, the output still forms a plausible shape.
On the contrary, with HRMs, there may be only one or several landmarks that are not robustly detected whereas the others are. 
In addition, they may be located at completely unreasonable positions according to the general morphology of the face.

\par
This is a well-known problem. Tai et al.~\cite{tai2019towards} proposed to improve the robustness by enforcing some temporal consistency. And the approach of Liu et al.~\cite{liu2019semantic} tries to correct the outliers by integrating a Coordinate Regression CNN at the end. 
Recently, Zou et al.~\cite{zou2019learning} introduced Hierarchical Structured Landmark Ensemble to impose additional structural constraints on HRMs. 
In these methods, the constraints are imposed in a post-processing step, which is not integrated into the HRM itself.
Therefore, all these methods either add complexity to the models or require learning on a video stream. 

\par

We propose a more general approach regularizing the output shape of HRMs by imposing additional geometric and global contextual constraints during training, directly integrated into the loss function. This adds no complexity during inference and can be trained on both image and video datasets. With exactly the same model structure, our models can effortlessly substitute the existing ones.\par

\textbf{Problem of current evaluation metrics for robustness: }
The most common metric for robustness is Failure Rate (FR). It measures the proportion of images in a (validation) set whose error is greater than a threshold. 
Table~\ref{Tab:dataset} shows the FR with an error threshold of 0.1 (FR$_{0.1}$) of HRNet. 
We find that this widely used FR$_{0.1}$ measure is almost ``saturated'' on several benchmarks such as COFW~\cite{burgos2013robust}, 300W~\cite{sagonas2013300}, 300W-Test and AFLW~\cite{koestinger11a}. That is, there are only 1 , 3 , 1 and 2 failure images respectively (bold numbers in Tab.~\ref{Tab:dataset}). 
This means that there are only very few challenging images for the state-of-the-art model HRNet in these datasets.
At this level, this indicator is saturated and becomes difficult to interpret when comparing the robustness of different methods as it is sensitive to random statistical variations. 
Therefore, it becomes necessary to modifiy the current evaluation metrics on these datasets and to find more challenging evaluation protocols to further decrease the gap with real-world application settings.
\par

\begin{table}[t]
\centering
\begin{tabular}{ccccccc}
\hline
 & COFW & 300W & 300W-Test & AFLW & WFLW & 300VW \\ \hline
Num. Landmarks & 29 & 68 & 68 & 19 & 98 & 68\\
Num. Train Images & 1,345 & 3,148 & / & 20,000 & 7,500 & 95,192\\
Num. Valid Images & 507 & 689 & 600 & 4,386 & 2,500 & 124,916\\ \hline
HRNet FR$_{0.1}$ (\%) & \textbf{0.19} & \textbf{0.44} & \textbf{0.33} & \textbf{0.046} & 3.12 & 2.26\\
FR (\%) per Image & \textbf{0.19} & \textbf{0.15} & \textbf{0.33} & \textbf{0.023} & 0.040 & $8.0\times10^{-4}$ \\
FR (\%) per Landmark & 0.0068 & 0.0021 & 0.0025 & 0.0012 & $4.1\times10^{-4}$ & $1.1\times10^{-5}$\\ \hline
\end{tabular}
\caption{Numerical details of the facial landmark datasets and the Failure Rate (FR) of HRNet on each dataset.}
\label{Tab:dataset}
\vspace{-8mm}
\end{table}

\section{Proposed evaluation metrics}
\label{sec:prop_metrics}
\textbf{Dataset:} The dataset is crucial to evaluate the robustness of the model. 
The most common robustness issues treated in the literature concern partial occlusions and large pose variations. 
COFW~\cite{burgos2013robust} is one of the first datasets that aims at benchmarking the performance of facial landmark detection under partial occlusion. 
300W~\cite{sagonas2013300} comprises a challenging validation subset with face images with large head pose variations, heavy occlusion, low resolution and complex lighting conditions. 
AFLW~\cite{koestinger11a} is a large-scale dataset including face images in extreme poses. 
WFLW~\cite{wayne2018lab} is a recently released dataset with even more challenging images. 
All the images are annotated in a dense format (98 points). 
The validation set of WFLW is further divided into 6 subsets based on the different difficulties such as occlusion, large pose or extreme expressions. 
300VW~\cite{shen2015first} is a video dataset annotated in the same format as 300W. 
The validation dataset is split into three scenarios, where the third one (300VW-S3) contains the videos in highly challenging conditions. 
\par
\textbf{Current Evaluation metrics: } The main performance indicator for facial landmark detection is the Normalised Mean Error:  $\mathrm{NME} = \frac{1}{N} \sum_i \mathrm{NME}_i$, an average over all $N$ images of a validation set, where for one image $i$ the error is averaged over all $M$ landmarks:
\begin{equation}
\mathrm{NME}_i =\frac{1}{M} \sum_{j} \mathrm{NME}_{i,j}   \; ,
\end{equation}
and for each landmark $j$:
\begin{equation}
\mathrm{NME}_{i,j} = \frac{\left \| \mathbf{S_{i,j}} - \mathbf{S_{i,j}^{*}} \right \|_{2}}{d_i} \; ,
\end{equation}
where $\mathbf{S_{i,j}, \mathbf{S}_{i,j}^{*}}\in \mathbb{R}^{2}$ denote the $j$-th predicted and the ground truth landmarks respectively.
For each image, we consider the inter-occular distance as normalisation distance $d_i$ for 300W, 300VW, COFW, WFLW and the face bounding box width for AFLW.\par

As mentioned before, Failure Rate FR$_\theta$ measures the proportion of the images in the validation set whose NME$_i$ is greater than a threshold $\theta$. 
We will denote this classical failure rate: FR$^I$ in the following.
In the literature, FR$^I_{0.1}$ and FR$^I_{0.08}$ are the principle metrics to measure the prediction robustness as they focus on rather large errors (i.e.\ 8\%/10\% of the normalisation distance).
\par
It is also very common to compute the FR$^I_\theta$ over the entire range of $\theta$, called the Cumulative Error Distribution (CED), which gives an overall idea on the distribution of errors over a given dataset.
Finally, for easier quantitative comparison of the performance of different models, the total area under the CED distribution can be computed, which is usually denoted as the Area Under Curve (AUC).
\par

We propose three modifications to these measures:\par

\textbf{Landmark-wise FR:} 
Instead of computing the average failure rate per image: FR$^I$, we propose to compute this measure \emph{per landmark}. That is, for each landmark $j$, the proportion of NME$_{i,j}$ larger than a threshold is determined. Finally, an average over all landmarks is computed, called FR$^L$ in the following.
There are two advantages of computing the failure rate in this way: 
(1) With HRMs, it happens that only one or few landmarks are not well detected (outliers). 
However, the FR$^I$ (\emph{per image}) may still be small because the rest of the landmarks are predicted with high precision and an average is computed per image. Thus, possible robustness problems of some individual landmarks are not revealed by the FR$^I$ measure.
(2) FR$^{L}$ can provide a finer granularity for model comparison, which is notably beneficial when the state-of-the-art methods have an FR$^I$ that is very close and almost zero on several benchmark datasets (see FR (\%) per Image/Landmark in Tab.~\ref{Tab:dataset}).

\par

\textbf{Cross-dataset validation:} Leveraging several datasets simultaneously is not new and has already been adopted by some previous works~\cite{smith2014collaborative,zhu2014transferring,zhang2015leveraging,wu2017leveraging,wayne2018lab}. 
Most of them focus on unifying the different semantic meanings among different annotation formats. 
In~\cite{zhu2019robust}, the authors validated the robustness of their model by training on 300W and validating on the COFW dataset.\par

We assume the reason why the performance of HRNet has ``saturated'' on several datasets is that the data distributions in the training and validation subsets are very close. Therefore, to effectively validate the robustness of a model, we propose to train it on a small dataset and test on a different dataset with more images to avoid any over-fitting to a specific dataset distribution. 
Thus, two important aspects of robustness are better evaluated in this way: firstly, the \emph{number} of possible test cases, which reduces the possibility to ``miss out'' more rare real-world situations. And secondly, the generalisation capacity to different data distributions, for example corresponding to varying application scenarios, acquisition settings etc.
\par


We propose four cross-dataset validation protocols: 
COFW$\rightarrow$AFLW (trained on COFW training set, validated on AFLW validation set with 19 landmarks), 
300W$\rightarrow$300VW, 
300W$\rightarrow$WFLW, 
and WFLW$\rightarrow$300VW. 
The annotation of 300W and 300VW has identical semantic meaning. 
On the other three protocols, we only measure the errors on the common landmarks between two formats. 
There are indeed slight semantic differences on certain landmarks.
However, in our comparing study this effect is negligible because: 
(1) We mainly focus on the large errors when validating the robustness.
That is, these differences are too small to influence the used indicators such as FR$^{L}_{0.1}$.
(2) When applying the same protocol for each compared model, this systematic error is roughly the same for all models.

\begin{figure}[t]
\centering
\begin{center}
\includegraphics[width=0.7\textwidth]{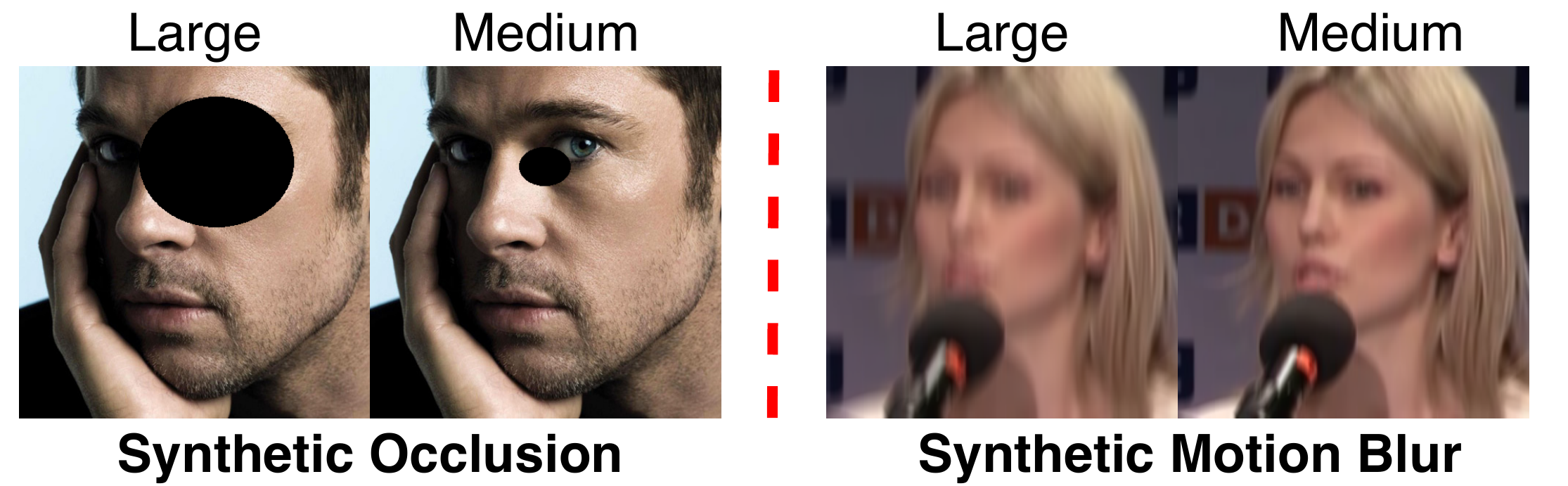}
\end{center}
\caption{An illustration of synthetic occlusion and motion blur protocol.}
\label{fig:synthetic occlusion}
\end{figure}

\textbf{Synthetic occlusion and motion blur:} Occlusion and motion blur are big challenges for robust facial landmark detection.
However, annotating the ground truth landmark positions of occluded/blurred faces is very difficult in practice. 
To further evaluate the robustness of the model against these noises, we thus propose to apply synthetic occlusions and motion blur on the validation images. 
For occlusion, a black ellipse of random size is superposed on each image at random positions.
For motion blur, inspired by~\cite{Sun_2019_FAB}, we artificially blur the 300VW dataset. 
For each frame at time $t$, the blurring direction is based on the movement of the nose tip (the 34th landmark) between the frame $t-1$ and $t+1$.
We adopt two protocols for both perturbations: large and medium, illustrated in Fig.~\ref{fig:synthetic occlusion}.
Obviously, the landmark detection performance of a model is deteriorated by these noises. 
But more robust models should be more resilient to these noise.
\par

\section{Proposed method}
\label{sec:prop_method}

We propose to add geometric and global constraints during the training of HRMs. Our method consists of the following three parts: 

\textbf{2D Wasserstein Loss:}
Sun et al.~\cite{sun2018integral} discussed the use of different loss functions for HRM. The most widely used loss function is heatmap $L2$ loss. It simply calculates the $L2$ norm of the pixel-wise value difference between the ground truth heatmap and the predicted heatmap.  
\par
We propose to train HRMs using a loss function based on the Wasserstein distance. Given two distributions $u$ and $v$ defined on $M$, the first Wasserstein distance between $u$ and $v$ is defined as:
\begin{equation}
    l_{1}(u, v)=\inf _{\pi \in \Gamma(u, v)} \int_{M \times M}|x-y| \mathrm{d} \pi(x, y),
\end{equation}
where $\Gamma(u, v)$ denotes the set of all joint distributions on $M \times M$ whose marginals are $u$ and $v$. The set $\Gamma(u, v)$ is also called the set of all couplings of $u$ and $v$. Each coupling $\pi(x, y)$ indicates how much “mass”
must be transported from the position $x$ to the position $y$ in order to transform the distributions $u$ into the distribution $v$.
\par
Intuitively, the Wasserstein distance can be seen as the minimum amount of “work” required to transform $u$ into $v$, where “work” is measured as the amount of distribution weight that must be moved, multiplied by the distance it has to be moved. 
This notion of distance provides additional geometric information that cannot be expressed with the point-wise $L2$ distance (see Fig.~\ref{fig:wloss1D}).
 \par

\begin{figure}[t]
\centering
\begin{subfigure}{\columnwidth}
  \centering
  \includegraphics[width=0.7\textwidth]{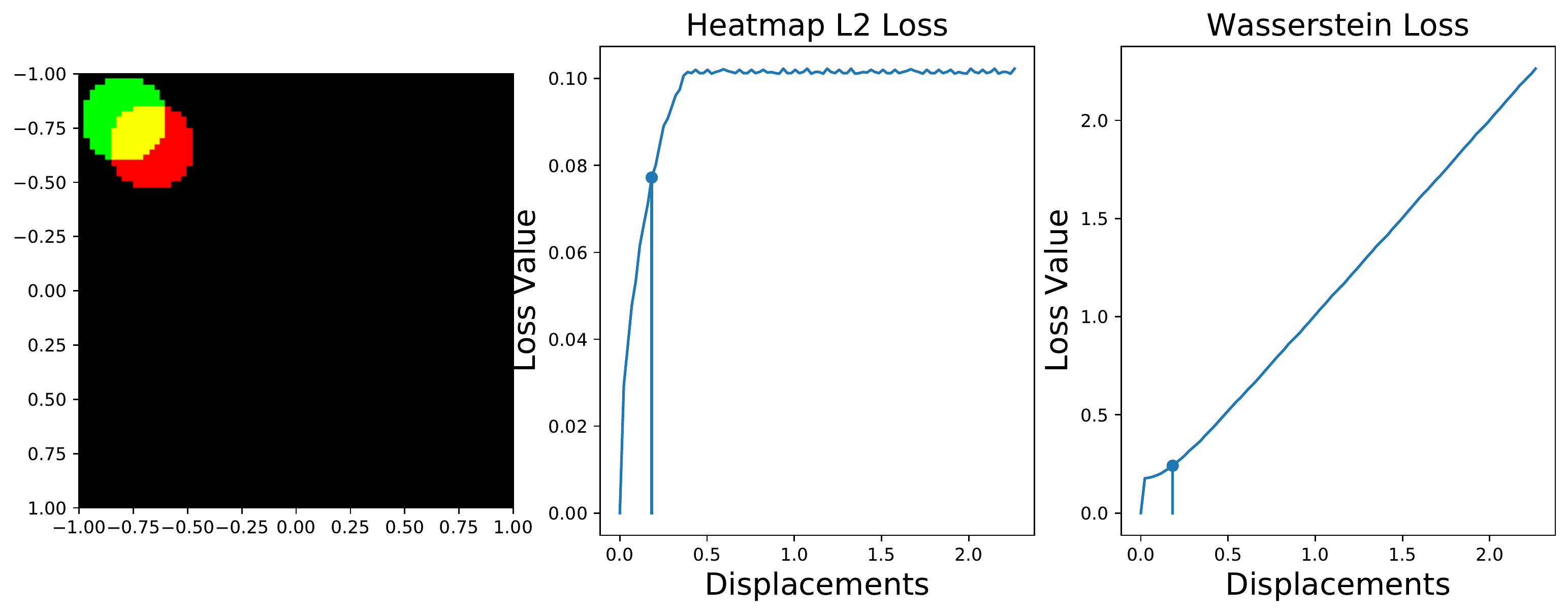}
  \caption{Ground truth (green) and predicted (red) distributions \textbf{overlap}.}
\end{subfigure}

\begin{subfigure}{\columnwidth}
  \centering
  \includegraphics[width=0.7\textwidth]{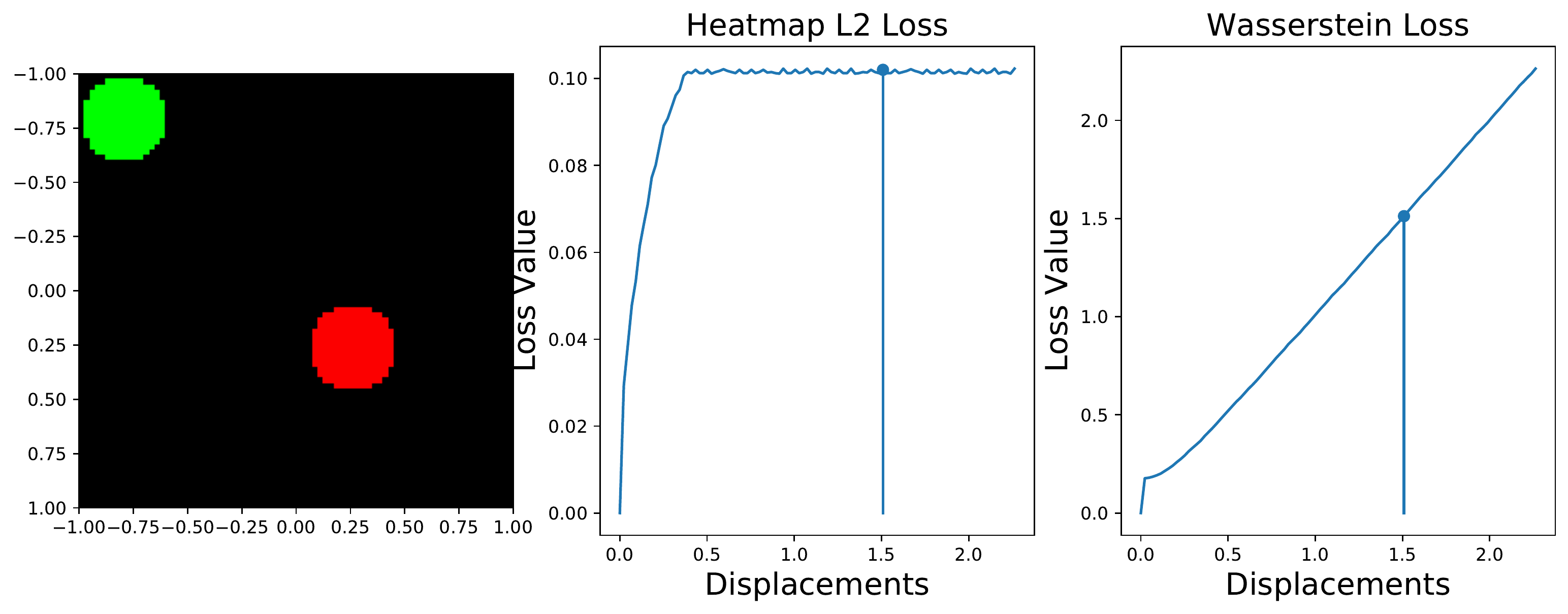}
  \caption{Ground truth and predicted distributions \textbf{do not overlap}.}
\end{subfigure}
\caption{Comparison of heatmap $L2$ loss and Wasserstein loss on 2D distributions. We observe that the value of $L2$ loss saturates when the two distributions do not overlap. However, the value of Wasserstein Loss continues to increase. The Wasserstein loss is able to better integrate the global geometry on the overall heatmap. (Figure taken from~\cite{Tralie2018} with slight modifications.)}
\label{fig:wloss2D}
\vspace{-0.4cm}
\end{figure}

To define our Wasserstein loss function for heatmap regression, we formulate the continuous first Wasserstein metric for two discrete 2D distributions $u',v'$ representing a predicted and ground truth heatmap respectively:
\begin{equation}
  L_W(u,v) = \min_{\pi' \in \Gamma'(u,v)} \sum_{x,y} \left| x-y \right|_2 \pi'(x,y) \,
\end{equation}
where $\Gamma'(u,v)$ is the set of all possible 4D distributions whose 2D marginals are our heatmaps $u$ and $v$, and $\left| \cdot \right|_2$ is the Euclidean distance.
The calculation of the Wasserstein distance is usually solved by linear programming and considered to be time-consuming. 
Previous work on visual tracking have developed differential Wasserstein Distance~\cite{zhao2008differential} and iterative Wasserstein Distance~\cite{yao2018visual} to boost the computation.
Cuturi~\cite{cuturi2013sinkhorn} proposed to add an entropic regularization and calculate an approximation of the loss by Sinkhorn iteration. 
This drastically accelerates the calculation and enables the gradient back-propagation through the loss calculation. 
Further, in our case, having discrete 2D distributions of size $64^2$ leading to a joint size of $64^4 \approx 1.67 \times 10^7$ (for ``weights'' and distances) as well as existing GPU implementations~\cite{viehmann2019implementation,Daza2019} make the computation of Wasserstein distance tractable.
A visual comparison of Wasserstein Loss and heatmap $L2$ loss on 2D distribution is presented in Fig.~\ref{fig:wloss2D}.

\par

Using Wasserstein loss for HRM has two advantages: 
(1) It makes the regression sensitive to the global geometry, thus effectively penalizing predicted activations that appear far away from the ground truth position. 
(2) When training with the $L2$ loss, the heatmap is not strictly considered as a distribution as no normalisation applied over the map. 
When training with the Wasserstein loss, the heatmaps are first passed through a softmax function. 
That means the sum of all pixel values of an output heatmap is normalised to 1, which is statistically more meaningful as each normalised value represents the probability of a landmark being at the given position. 
Moreover, when passed through a softmax function, the pixel values on a heatmap are projected to the $e$-polynomial space. 
This highlights the largest pixel value and suppresses other 
pixels whose values are inferior.
\par

\begin{figure}[t]
  \raggedleft
  \begin{minipage}{0.49\textwidth}
    \includegraphics[width=0.9\linewidth]{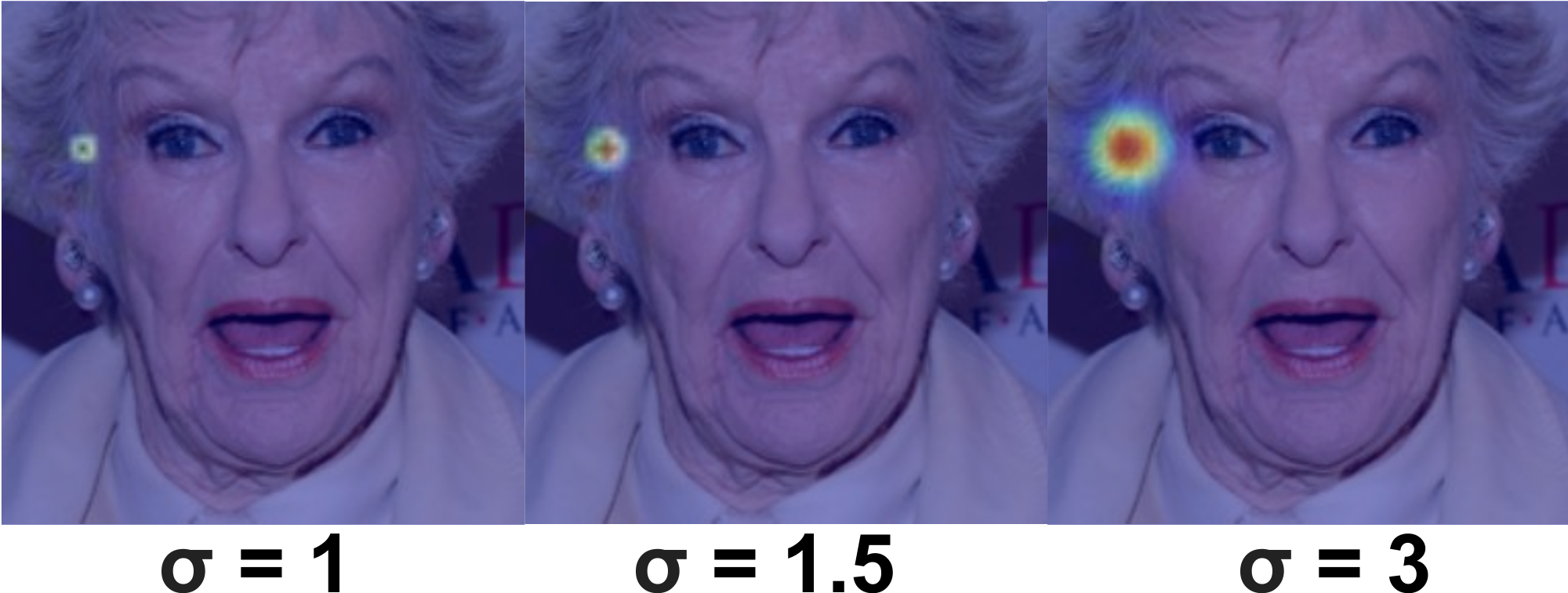}
    \caption{Illustration of ground truth target heatmaps defined by Gaussian functions with different $\sigma$.}
    \label{fig:sigma}
  \end{minipage}
  \raggedright
  \begin{minipage}{0.49\textwidth}
    \raggedright
    \includegraphics[width=0.9\linewidth]{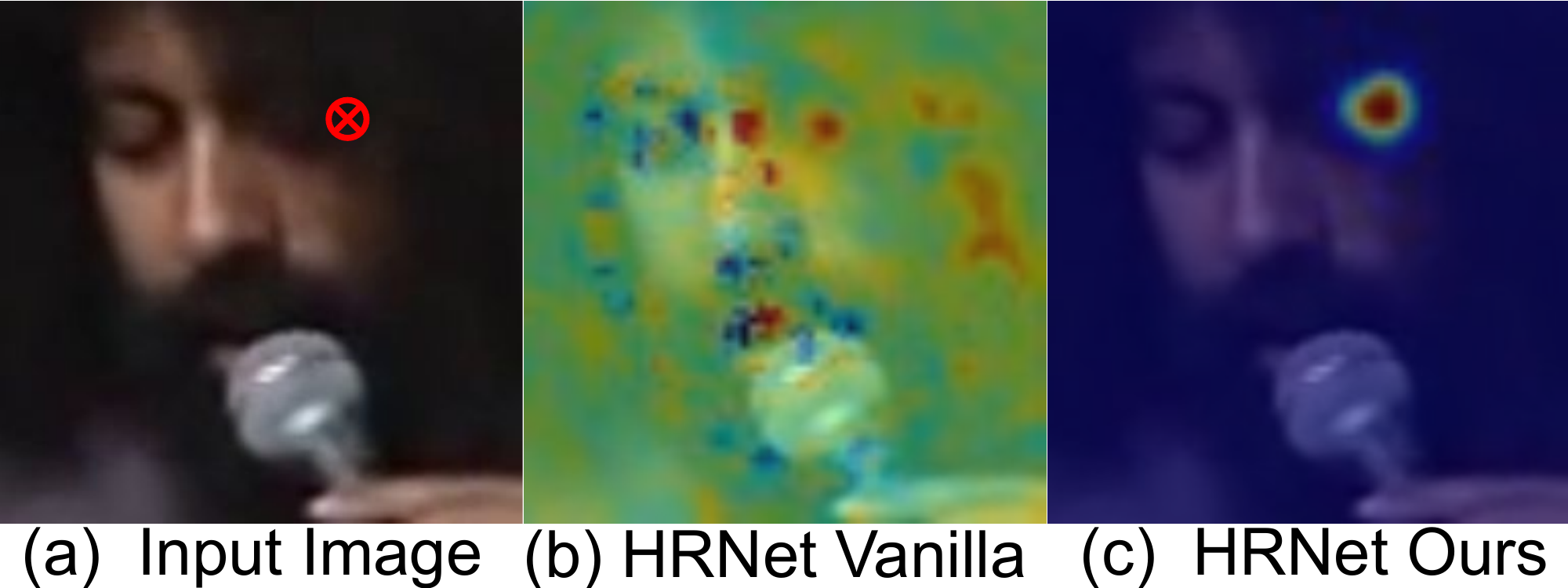}
    \caption{Output heatmap comparisons under occlusion. We show the heatmaps of the landmark marked in red.}
\label{fig:hm_comp}
  \end{minipage}
\end{figure}

\textbf{Smoother target heatmaps:}
To improve convergence and robustness, the values of the ground truth heatmaps of HRMs for facial landmark detection are generally defined by 2D Gaussian functions, where the parameter $\sigma$ is commonly set to 1 or 1.5 (see Fig.~\ref{fig:sigma}).

Intuitively, enlarging $\sigma$ will implicitly force the HRM to consider a larger local neighborhood in the visual support throughout the different CNN layers. 
Therefore, when confronting partial interferences (e.g.~occlusion, bad lighting conditions), the model should consider a larger context and thus be more robust to these types of noise.  
Nonetheless, the Gaussian distribution should not be too spread out to ensure some precision and to avoid touching
the map boundaries.

We empirically found that $\sigma=3$ is an appropriate setting for facial landmark detection. 
In our experiments, we systematically demonstrate the effectiveness of using $\sigma=3$ compared to $\sigma=1$ or $\sigma=1.5$ for robust landmark detection under challenging conditions.  

\par

\textbf{Predicted landmark sampling:}
In the early work of HRM~\cite{newell2016stacked,bulat2017far}, the position of a predicted landmark $p$ is sampled directly at the position of the maximum value of the given heatmap $H$:
\begin{equation}
    (p_{x},p_{y}) = \mathop{\arg\max}_{p}(H).
\end{equation}
However, this inevitably leads to considerable quantization error because the size of the heatmap is generally smaller than the original image (usually around 4 times). 
An improvement is to use interpolation and resample the numerical coordinates using 4 neighbouring pixel (bilinear interpolation). We denote this method as ``GET\_MAX''.
\par

Liu et al. discussed in~\cite{liu2019semantic} that using a target Gaussian distribution with bigger $\sigma$ decreases the overall NME. 
Indeed, using bigger $\sigma$ flattens the output distribution and therefore obfuscates the position of the peak value. As a result, the predictions are locally less precise.
\par

To compensate this local imprecision when using bigger $\sigma$, we propose another approach to sample numerical coordinates from the heatmap. 
Inspired by~\cite{sun2018integral}, we propose to use the spatial barycenter of the heatmap:
\begin{equation}
    (p_{x},p_{y}) = \int_{q \in \Omega} q\cdot H(q) \; ,
\end{equation}
where $\Omega$ denotes the set of pixel positions on the heatmap. 
We denote this method as ``GET\_BC''~(BaryCenter). 
\par

GET\_BC enables sub-pixel prediction, which effectively improves the local precision of the model trained with Wasserstein loss and big $\sigma$. 
On the other hand, GET\_BC considers the entire heatmap and thus involves a global context for a more robust final detection.
\par

\par

\section{Experiments}
\label{sec:exp}
In this section, we compare our method with other state-of-the-art methods and realize ablation studies using both traditional and proposed evaluation metrics. 
We also apply our method on various HRMs to demonstrate that our method can be directly used for any structures without further adjustments.

\begin{table}[t]
\centering
\begin{tabular}{l||llll|llll}
\hline
Sampling Method & $\sigma$ & Loss&  NME (\%) & FR$^{L}_{0.2}$ (\%) & $\sigma$ & Loss& NME (\%) & FR$^{L}_{0.2}$ (\%)\\ \hline
GET\_MAX & $1$&HM $L2$ &  \textbf{3.34} & \textbf{0.58} & $3$&W Loss & 4.00 & 0.63 \\
GET\_BC & & & 20.15 & 16.83 & &  & \textbf{3.46} & \textbf{0.54}  \\ \hline
\end{tabular}
\caption{Performance of HRNet on the 300W validation set when using  different coordinate sampling methods. GET\_BC improves the local precision (see FR$^{L}_{0.05}$) of the model trained with Wasserstein loss (W Loss) and large $\sigma$. However, it harms the performance of the model trained with Heatmap $L2$ loss (HM $L2$).}
\label{Tab:get_BC}

\end{table}

\begin{figure}[!t]
\begin{minipage}[t]{\textwidth}
  \centering
  \begin{minipage}{0.41\textwidth}
    \centering
    \includegraphics[width=0.835\textwidth]{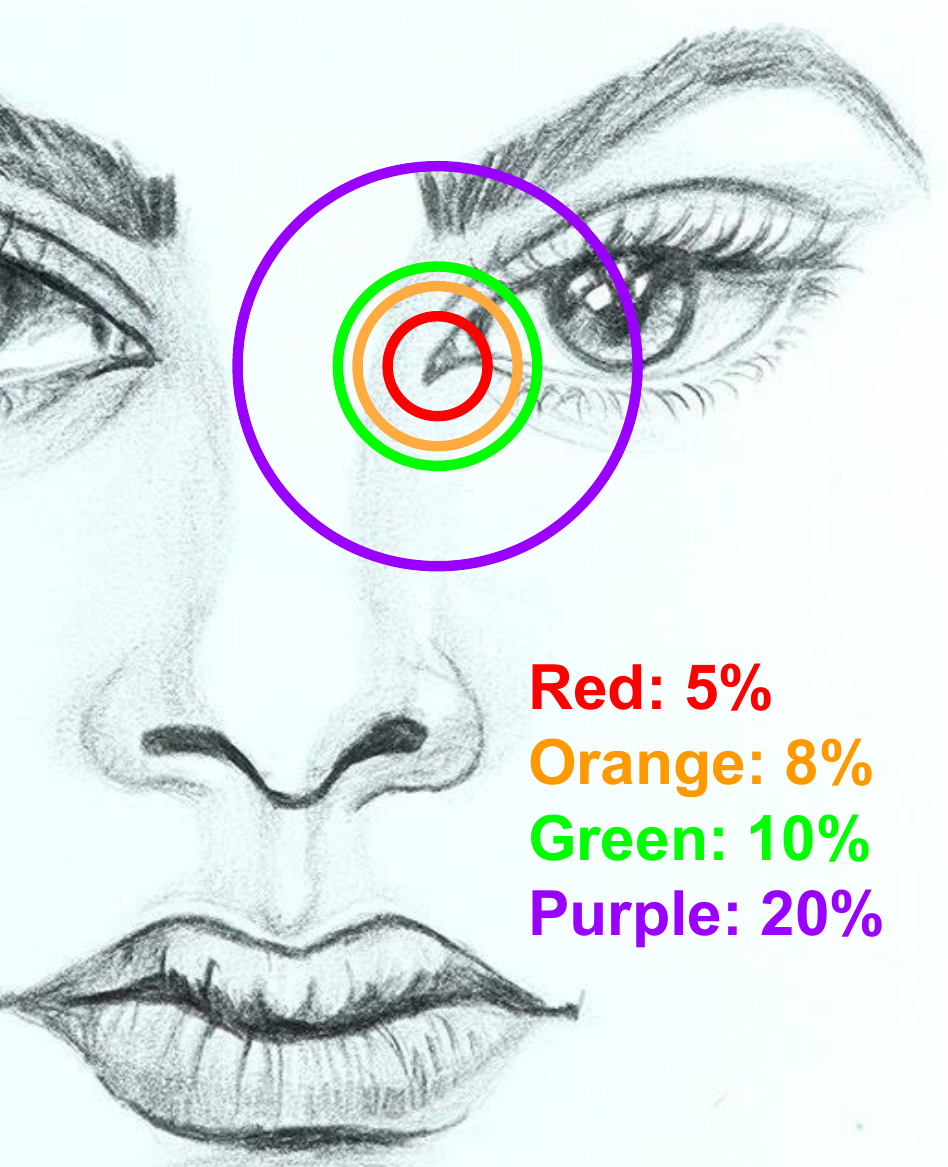}
    \captionof{figure}{Demonstration of the normalised error from 5\% to 20\%.}
    \label{Fig:error_percentage}
  \end{minipage}
  \hfill
  \centering
  \begin{minipage}{0.55\textwidth}
    \centering
        \begin{tabular}{llll}
        \hline
        Method & S1 & S2 & S3 \\ \hline
        TSTN~\cite{liu2017two} & 5.36 & 4.51 & 12.84 \\
        DSRN~\cite{miao2018direct} & 5.33 & 4.92 & 8.85 \\
        FHR+STA~\cite{tai2019towards} & 4.42 & 4.18 & 5.98 \\
        SA~\cite{liu2019semantic} & 3.85 & \textbf{3.46} & 7.51 \\
        FAB~\cite{Sun_2019_FAB} & \textbf{3.56} & 3.88 & \textbf{5.02} \\
        DeCaFA~\cite{Dapogny_2019_decafa} & 3.82 & 3.63 & 6.67 \\\hline \hline
        HRNet &  &  &  \\ \hline
        $\sigma=1$, $L2$ & 3.74 & 3.73 & 5.49 \\
        $\sigma=3$, $L2$ & 3.42 & \textbf{3.58} & 5.12 \\
        $\sigma=1$, W Loss & 3.41 & 3.66 & 5.01 \\
        $\sigma=3$, W Loss & 3.39 & 3.64 & \textbf{4.99} \\
        $\sigma=3$, W Loss, CC & \textbf{3.35} & 3.61 & 5.05 \\ \hline
        \end{tabular}
      \captionof{table}{NME (\%) on 300VW. W~Loss - Wasserstein Loss. CC - CoordConv.}
      \label{Tab:300vw_valid_conventional}
    \end{minipage}
\end{minipage}
\end{figure}

To provide a general idea on the NME and the threshold of FR in this section, we demonstrate the error normalised by inter-ocluar distance at different scales in Fig.~\ref{Fig:error_percentage}. 
The ground truth position is the inner corner of the right eye. 
The errors within 5\% are relatively small ones. 
From 10\%, the errors might be larger than the distance between adjacent landmarks. 
The errors larger than 20\% completely violate the reasonable face shape and needs to be avoided in most applications.
\par

\textbf{Effectiveness of barycenter sampling:} 
The GET\_BC method for estimating the predicted landmark coordinates is able to significantly improve the precision of the model trained with Wasserstein loss and larger $\sigma$ (see Tab.~\ref{Tab:get_BC}, NME is improved from 4.00\% to 3.46\%).

In contrast, GET\_BC is not compatible with the output trained with heatmap $L2$ loss (FR$^{L}_{0.2}$ is largely increased from 0.58\% to 16.83\% using GET\_BC). 
Training with $L2$ is less robust and generally leads to spurious activations far away from the ground truth position, which prevents GET\_BC from estimating good positions.
Figure~\ref{fig:hm_comp} shows an example comparing the output heatmaps from a vanilla HRNet (trained with $L2$ loss, $\sigma=1$) and our HRNet (trained with Wasserstein loss, $\sigma=3$) on a occluded landmark (outer right eye-corner).
We observe that our strategy effectively removes the spurious activation on the unrelated regions, so that the prediction will be more robust and GET\_BC can be effective. 
Therefore, in the following experiments, we will by default use GET\_MAX for models trained with the $L2$ loss and GET\_BC for models trained with the Wasserstein loss.
\par

\textbf{Comparison with the state-of-the-art:} 
We performed an ablation study using a ``vanilla'' HRNet (trained with heatmap $L2$ loss and $\sigma=1$) as our baseline. 
We also tested a recent method called CoordConv (CC)~\cite{liu2018intriguing} to integrate geometric information to the CNN. To this end, we replaced all the convolutional layers by CoordConv layers.
First, we benchmark our method with standard evaluation metrics NME on 300VW in Tab.~\ref{Tab:300vw_valid_conventional} and WFLW in Tab.~\ref{Tab:wflw_valid_conventional}. 
As mentioned in Sect.~\ref{sec:emp_study}, the performance of vanilla HRNet on 300W, AFLW and COFW has already reached a high level. 
Thus, there are only very few challenging validation images for HRNet.
In this case, the NME is dominated by a large amount of small errors, and it can thus no longer reflect the robustness of the models.
Therefore, we put the comparison using standard metrics on 300W, AFLW and COFW in the supplementary materials.
In the paper, we will demonstrate the robustness of the models trained on 300W, AFLW and COFW by using cross-dataset validation in the following experiments. 
\par

On 300VW (Tab.~\ref{Tab:300vw_valid_conventional}), our method shows promising performance, especially under challenging conditions on Scenario 3. 
On S3, by using Wasserstein loss, the NME drops by 0.48 point. 
By using a bigger $\sigma$, the NME drops by 0.37 point.
By using both, the NME can be further improved for a small margin. 
Using the Wasserstein loss combined with a larger $\sigma$, our method outperforms the vanilla HRNet by a significant margin of 0.39\%, 0.15\% and 0.5\% points on scenario 1, 2 and 3 respectively, and outperforms state-of-the-art methods.
\par

On WFLW (Tab.~\ref{Tab:wflw_valid_conventional}), our method achieves good performances by using a strong baseline. Nonetheless, using Wasserstein loss (4.57\%) only achieves marginal improvement compared to using $L2$ loss (4.60\%). 
We think that it is because the predictions are already ``regularized'' by the dense annotation of WFLW. We will analyze this issue in detail in Sect.~\ref{sec:discussion}.
\par

\begin{table}[t]
\begin{tabular}{c|c|c|c|c|c|c}
\hline
Method & ESR~\cite{cao2014face} & SDM~\cite{xiong2013supervised} & CFSS~\cite{zhu2015face} & DVLN~\cite{wu2017leveraging} & Wing~\cite{feng2018wing} & LAB~\cite{wayne2018lab} \\ \hline
WFLW  & 11.13 & 10.29 & 9.07 & 6.08 & \textbf{5.11} & 5.27 \\ \hline \hline
HRNet & $\sigma=1.5$, $L2$  & $\sigma=3$, $L2$ & $\sigma=1.5$, W & $\sigma=3$, W & $\sigma=1.5$, W, CC  & $\sigma=3$, W, CC \\ \hline
WFLW & 4.60 & 4.76 & 4.57 & 4.73 & \textbf{4.52} & 4.82 \\ \hline
\end{tabular}
\caption{NME (\%) on WFLW. W - Wasserstein Loss. CC - CoordConv.}
\label{Tab:wflw_valid_conventional}
\vspace{-5mm}
\end{table}

\begin{figure}[!t]
\begin{minipage}[t]{\textwidth}
  \begin{minipage}{0.375\textwidth}
    \centering
    \includegraphics[width=\textwidth]{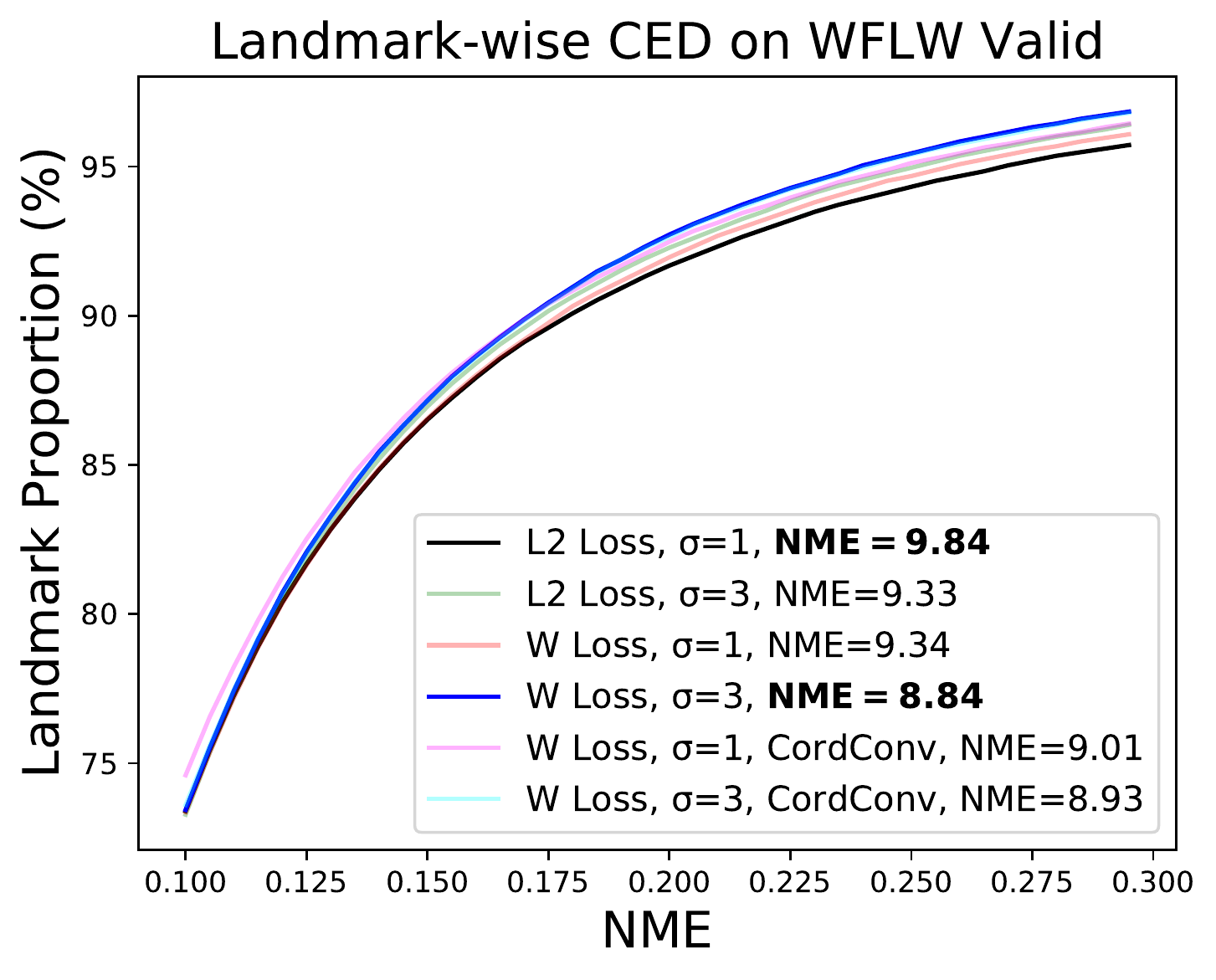}
    \captionof{figure}{Landmark-wise CED of 300W$\rightarrow$WFLW validation.}
    \label{fig:cross_300w}
  \end{minipage}
  \hfill
  \begin{minipage}{0.58\textwidth}
    \scriptsize
    \centering
        \begin{tabular}{lllllll}
            \hline
            Method & \multicolumn{2}{c}{Scenario 1} & \multicolumn{2}{c}{Scenario 2} & \multicolumn{2}{c}{Scenario 3} \\ 
            &NME&FR$^{L}_{0.1}$& NME & FR$^{L}_{0.1}$  & NME & FR$^{L}_{0.1}$ \\ \hline

            $\sigma=1$, $L2$ & 4.44& 5.02 & 4.37& 4.86 & 6.67 & 11.65\\
            $\sigma=3$, $L2$ & 4.36& 4.89 & 4.38 & 4.83 & 6.35& 10.97\\
            $\sigma=1$, W & 4.16& 4.68 & 4.21 & 4.67 & 6.31& 10.08\\
            $\sigma=3$, W & 4.17& 4.84 & 4.16 & 4.47 & \textbf{6.01}& 9.91\\
            $\sigma=1$, W, CC & \textbf{4.05} & \textbf{4.22} & \textbf{4.11}& \textbf{4.26} & 6.32& 10.61\\
            $\sigma=3$, W, CC & 4.21 & 4.78 & 4.24 & 4.61 & 6.02& \textbf{9.58}\\ \hline
        \end{tabular}
      \captionof{table}{NME (\%) and FR$^{L}_{0.1}$ (\%) comparision of 300W$\rightarrow$300VW cross-dataset validation using HRNet.  W - Wasserstein Loss. CC - CoordConv.}
      \label{Tab:cross_300w}
    \end{minipage}
\end{minipage}
\end{figure}

\textbf{Cross-dataset validation:} 
We use cross-dataset validation to measure the robustness of our HRNet trained on 300W.
We present the landmark-wise CEDs of protocol 300W$\rightarrow$WFLW (see Fig.~\ref{fig:cross_300w}) 
and protocol 300W$\rightarrow$300VW (see Tab.~\ref{Tab:cross_300w}). 

From Fig.~\ref{fig:cross_300w}, we find that using larger $\sigma$ (L2 Loss, $\sigma=3$) and Wasserstein loss (W Loss, $\sigma=1$) can respectively improve the NME by 0.5 point. 
Using both (W Loss, $\sigma=3$) further improves the NME to be 1 point inferior than the vanilla HRNet.
Notably, the improvement on larger errors (error $>=20\%$) is more significant the the errors $<20\%$, which demonstrates the superior robustness against large errors of our HRNet compared to vanilla HRNet.
From Tab.~\ref{Tab:cross_300w}, we obtain similar conclusions. 
Both bigger $\sigma$ and Wasserstein Loss improve the robustness.
The contribution of Wasserstein Loss is more important than the larger $\sigma$ (see FR$^{L}_{0.1}$).
However, even GET\_BC is used, a larger $\sigma$ still slightly decreases the local precision. 
As a result, on the less challenging datasets such as 300VW-S1 and 300VW-S2, we found that the best performance can be obtained by using a combination of small $\sigma$, Wasserstein loss and CoordConv.
On more challenging datasets such as WFLW and 300VW-S3, the best performance is obtained by using a combination of the Wasserstein loss and a larger $\sigma$.

Protocols COFW$\rightarrow$AFLW and WFLW$\rightarrow$300VW are presented in the supplementary materials.
Specifically, our method achieves a bigger improvement on COFW$\rightarrow$AFLW-All compared to COFW$\rightarrow$AFLW-Frontal. This is because AFLW-All contains non-frontal images, which is more challenging than AFLW-Frontal. 

\textbf{Synthetic occlusions and motion blur:} We further evaluated the robustness against synthetic perturbations that we described in Sect.~\ref{sec:prop_metrics} (see Tab.~\ref{Tab:synocc_300W}). 
We find that the model is more robust to occlusion and motion blur by using a larger $\sigma$ and Wasserstein loss.
For example, the FR$^{L}_{0.2}$ is improved from 2.66\% to 1.72\% under large occlusions.
Under large motion blur perturbations, the FR$^{L}_{0.2}$ is improved from 36.63\% to 31.32\%.
\par

\begin{table}[t]
\centering
\begin{tabular}{l|lll|lllll|lllll}
\hline
 \multicolumn{4}{c|}{}   & \multicolumn{5}{c|}{Occlusion (300W)} & \multicolumn{5}{c}{Blur (300VW-S3)}\\ \hline
Protocol & $\sigma$ & Loss & CC & NME & FR$^{I}_{0.1}$ & FR$^{L}_{0.1}$ & FR$^{L}_{0.15}$ & FR$^{L}_{0.2}$ & NME & FR$^{I}_{0.1}$ & FR$^{L}_{0.1}$ & FR$^{L}_{0.15}$ & FR$^{L}_{0.2}$\\ \hline
\multirow{3}{*}{Large} & 1 & L2 & $\times$ & 4.63 & 4.31 & 10.11 & 4.83 & 2.66 & 27.42 & 66.3 & 56.57 & 44.28 & 36.63 \\
 & 3 & W & $\times$ & \textbf{4.48} & \textbf{2.95} & \textbf{9.62} & \textbf{3.88} & \textbf{1.72} & \textbf{19.15} & 64.51 & \textbf{54.70} & \textbf{41.03} & \textbf{31.32} \\
 & 3 & W & \checkmark & 4.60 & 3.79 & 9.89 & 4.26 & 2.03 & 19.32 & \textbf{63.65} & 55.71 & 41.85 & 31.68 \\ \hline
\multirow{3}{*}{Medium} & 1 & L2 & $\times$ & \textbf{3.57} & 0.97 & 5.58 & 1.94 & 0.83 & 11.07 & 31.67 & 28.34 & 15.22 & 9.32 \\
 & 3 & W & $\times$ & 3.62 & \textbf{0.46} & 5.46 & 1.74 & 0.63 & 9.5 & 27.54 & 27.34 & 14.35 & 7.80 \\
 & 3 & W & \checkmark & 3.6 & 0.58 & \textbf{5.11} & \textbf{1.65} & \textbf{0.61} & \textbf{9.07} & \textbf{25.14} & \textbf{26.37} & \textbf{12.76} & \textbf{6.27}\\ \hline
\end{tabular}
\caption{Results of the HRNet with synthetic occlusion (validated on 300W dataset) and synthetic motion blur (validated on 300VW-S3). 
}
\label{Tab:synocc_300W}
\end{table}

\textbf{Comparision with other loss functions:} 
Besides Heatmap $L2$, we note that there exists several other loss functions for HRMs. Jensen–Shannon divergence (loss) is a common metric for measuring the distance between two probabilistic distributions. Soft ArgMax~\cite{sun2018integral} transforms the heatmap regression into a numeric integral regression problem, which we think might be beneficial for model robustness. From Tab.~\ref{Tab:loss}, we find that the HRNet trained with Wasserstein loss delivers more robust predictions compared to the HRNet trained with other loss functions.

\begin{table}[t]
\centering
\begin{tabular}{l|ll|llllll}
\hline
Dataset  & Loss & Sampling & FR$^{I}_{0.08}$ & FR$^{I}_{0.1}$ & FR$^{L}_{0.08}$ & FR$^{L}_{0.1}$ & FR$^{L}_{0.15}$ & FR$^{L}_{0.2}$ \\ \hline

\multirow{5}{*}{WFLW}  & Jensen–Shannon$^{\dagger}$ & GET\_MAX  & 40.48 & 26.24 & 37.60 & 26.89 & 13.72 & 8.37\\
 & Jensen–Shannon$^{\dagger}$ & GET\_BC   & 40.44 & 26.12 & 37.49 & 27.17 & 14.37 & 9.04\\
 & Soft ArgMax$^{\dagger}$ & GET\_BC  & 42.60 & 26.40 & 40.37 & 29.23 & 14.86 & 8.51\\
 & Wasserstein$^{\dagger}$ & GET\_BC  & \textbf{39.24} & 25.12 & \textbf{37.27} & 26.61 & 13.42 & 8.03\\
 & Wasserstein$^{*}$ & GET\_BC  & 39.96 & \textbf{23.64} & 37.42 & \textbf{26.39} & \textbf{12.81} & \textbf{7.32}\\ \hline
 
\multirow{4}{*}{300VW-S3}   & Jensen–Shannon$^{\dagger}$ & GET\_MAX   & 11.07 & 5.14 & 19.27 & 11.97 & 4.45 & 2.06 \\
 & Jensen–Shannon$^{\dagger}$ & GET\_BC  & 10.87 & 5.34 & 18.93 & 11.92 & 4.66 & 2.28\\
 & Soft ArgMax$^{\dagger}$ & GET\_BC  & 11.08 & 5.61 & 19.00 & 11.59 & 4.45 & 2.09\\
 & Wasserstein$^{\dagger}$ & GET\_BC  & 9.72 & 3.73 & 17.96 & 11.08 & 4.15 & 1.79\\
 & Wasserstein$^{*}$ & GET\_BC  & \textbf{7.52} & \textbf{2.96} & \textbf{16.03} &\textbf{9.58} & \textbf{3.39} & \textbf{1.46}\\\hline

\end{tabular}
\caption{Cross-dataset validation (300W$\rightarrow$WFLW \& 300W$\rightarrow$300VW) of the HRNet using different loss functions. $\dagger$: Trained with Gaussian Distribution $\sigma=1$ without CoordConv. $*$: Trained with $\sigma=3$ with CoordConv. }
\label{Tab:loss}
\end{table}

\begin{figure}[!t]
\begin{minipage}[t]{\textwidth}
  \begin{minipage}{0.62\textwidth}
  \captionsetup{type=figure}
    \centering
    \subcaptionbox{300W$\rightarrow$WFLW}
    {\includegraphics[width=0.475\textwidth]{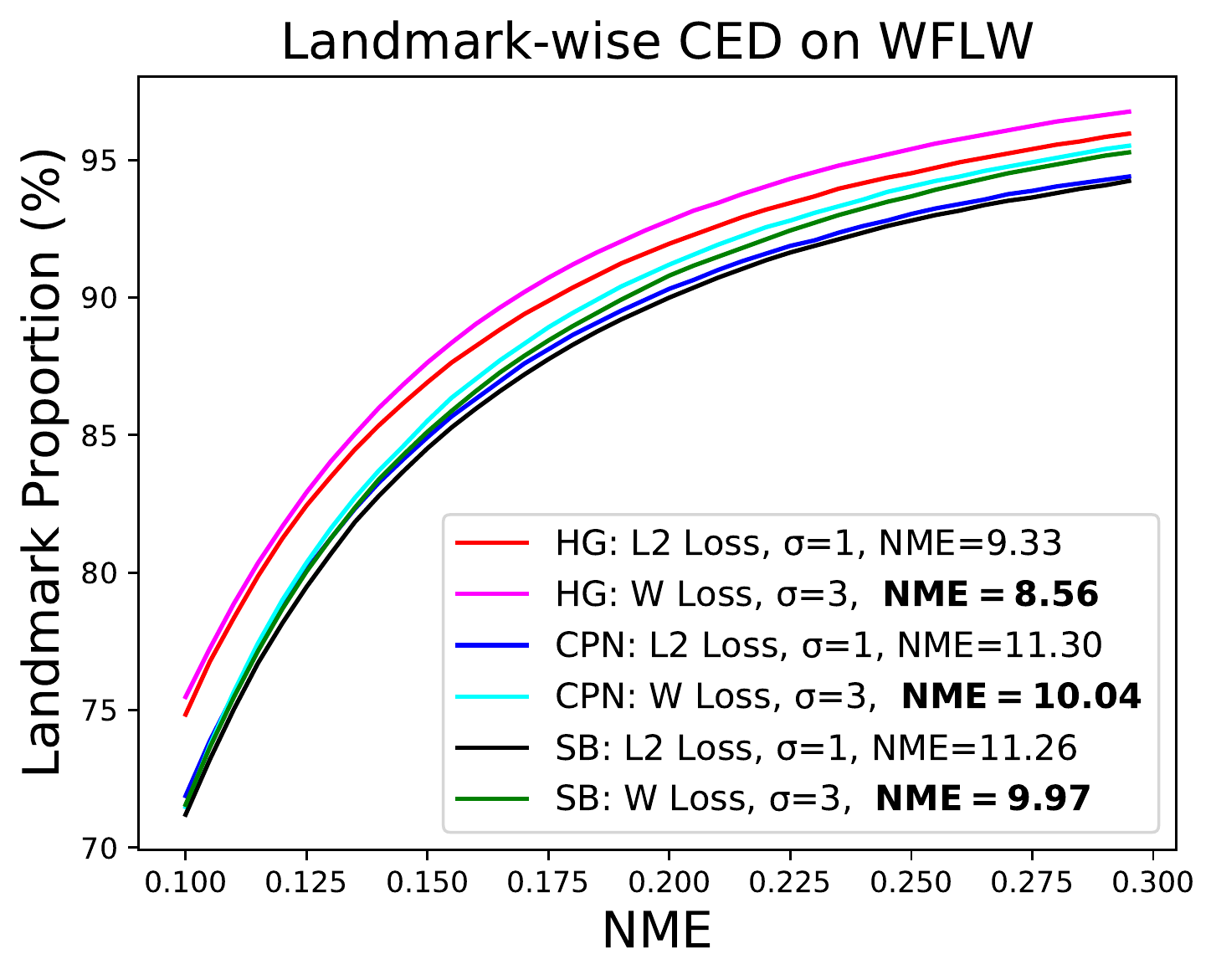}}\quad
    \subcaptionbox{300W$\rightarrow$300VW-S3}
     {\includegraphics[width=0.475\textwidth]{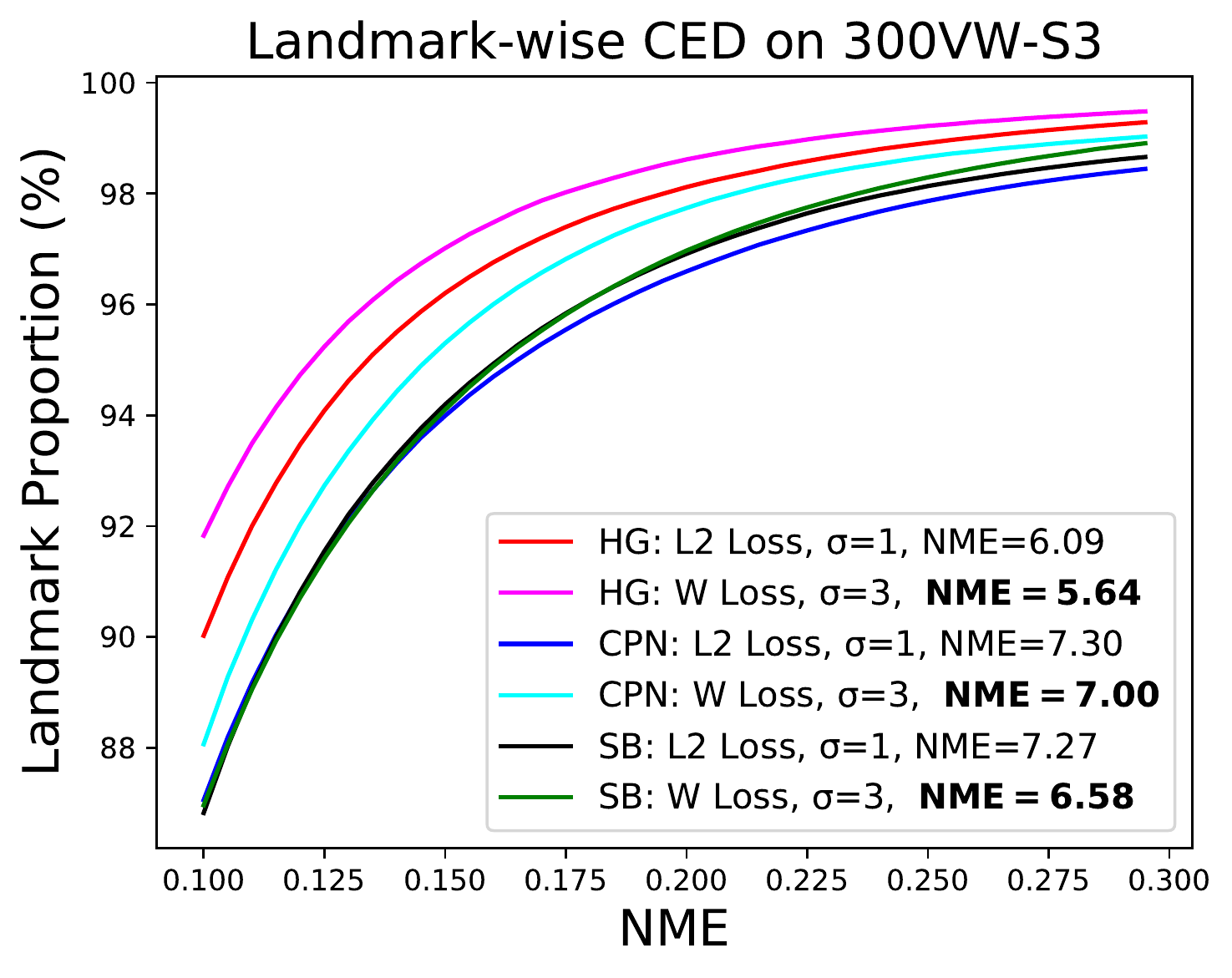}}
    \caption{Cross-dataset validation of HG~\cite{newell2016stacked}, CPN~\cite{chen2018cascaded} and SimpleBaselines(SB)~\cite{xiao2018simple}.}
    \label{fig:3models}
  \end{minipage}
  \hfill
  \begin{minipage}{0.37\textwidth}
    \scriptsize
    \centering
        \begin{tabular}{lll|ll}
            \hline
            N. Landmks  & $\sigma$ & Loss & FR$^{L}_{0.15}$ & FR$^{L}_{0.2}$ \\ \hline
            \multirow{2}{*}{17} & 1 & L2 & 2.79 & 1.60 \\
            & 3 & W & \textbf{2.68} & \textbf{1.29} \\ \hline
            \multirow{2}{*}{68} & 1 & L2 & 0.65 & 0.37 \\
             & 3 & W & \textbf{0.62} & \textbf{0.33} \\ \hline
            \multirow{2}{*}{98} & 1 & L2 & 0.44 & 0.25 \\
             & 3 & W & \textbf{0.43} & \textbf{0.22} \\ \hline
            \end{tabular}
      \captionof{table}{Comparison of HRNets trained with different number of landmarks on WFLW.  W: Wasserstein Loss.}
      \label{Tab:num_landmarks}
    \end{minipage}
\end{minipage}
\end{figure}

\textbf{Different models:} To demonstrate that our method can be used on a variety of HRMs regardless of the model structure, we test our method on three popular HRMs: HourGlass~\cite{newell2016stacked}, CPN~\cite{chen2018cascaded} and SimpleBaselines~\cite{xiao2018simple}. In Fig.~\ref{fig:3models} we can see that all of the three models benefit from our method. This indicates that our approach is quite general and can be applied to most existing HRMs.
\par

\begin{figure}[t]
\centering
\begin{center}
\includegraphics[width=\linewidth]{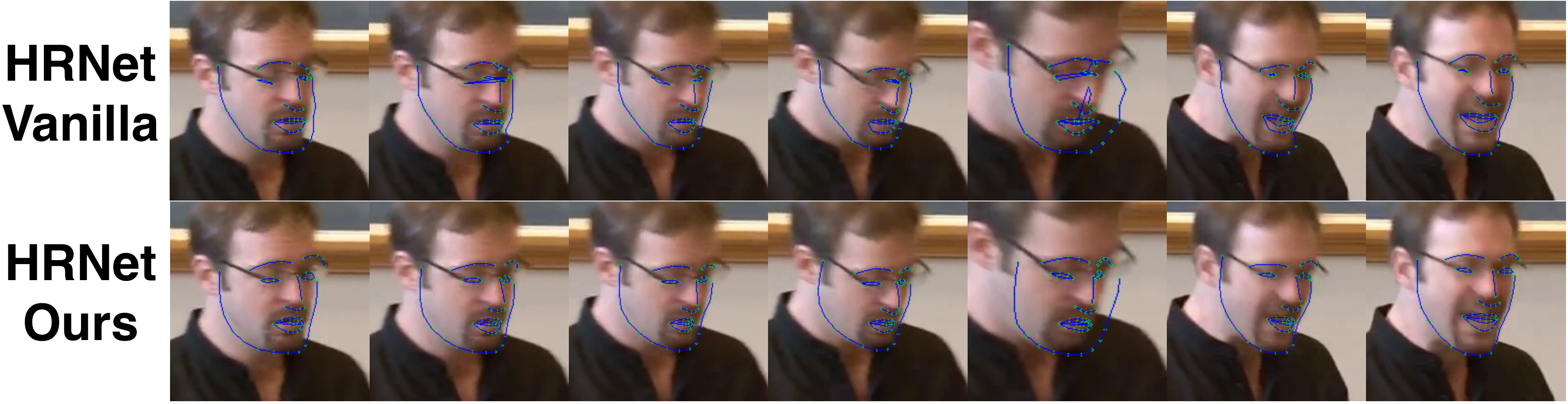}
\end{center}
\caption{Visual comparison of vanilla HRNet ($L2$ Loss and $\sigma=1$) and our HRNet (Wasserstein loss and $\sigma=3$).}
\label{fig:visual_comp}
\end{figure}

\textbf{Visual comparison:} We visually compare the predictions from vanilla HRNet and our HRNet on a challenging video clip in Fig.~\ref{fig:visual_comp}.
Our HRNet gives a more robust detection when confronted to extreme poses and motion blur. 
By using the Wasserstein loss, a larger $\sigma$ and GET\_BC, the predicted landmarks are more regularized by the global geometry compared to the prediction from the vanilla HRNet.
\par

\section{Discussions}
\label{sec:discussion}
\textbf{Does dense annotation naturally ensure the robustness?} We find that our method shows less significant improvement on the model trained on WFLW.
Intuitively, we presume that by training with a dense annotation (98 landmarks), the model predictions are somewhat regularized by the correlation between neighbouring landmarks. 
In Tab.~\ref{Tab:num_landmarks}, we compare the models trained with different number of landmarks.
The 68 landmark format is a subset of the original 98 landmark format, which is similar to the 300W annotation.
The 17 landmark format is a subset of the 68 landmark format, which is similar to the AFLW annotation (except the eye centers).
To ensure the fair comparison, though trained with different number of landmarks, all the models listed are tested on the common 17 landmarks.
We found that the prediction is naturally more robust by training with denser annotation formats. 
Therefore, compared to the model trained with more sparse annotation, our method achieves less important improvement on the model trained with dense annotation.
\par


\textbf{Recommended settings:} We recommend to use the Wasserstein loss and GET\_BC to improve the robustness of the model in all cases. 
Using a larger $\sigma$ will significantly improve the robustness under challenging conditions. 
Nonetheless, it deteriorates the local precision at the same time.
In fact, the value of $\sigma$ is a trade-off between robustness and precision.
Therefore, we recommend to use a larger $\sigma$ only when confronting crucial circumstances. 
When facing less challenging conditions, we recommend to use a combination of Wasserstein loss and small $\sigma$. 
Complementing CoordConv with Wasserstein loss and small $\sigma$ will further improve the NME performance.
However, it adds slight computational complexity to the HRMs. Specifically, when using small $\sigma$, the models with CoordConv are less robust against large occlusions compared to those without CoordConv.
\par

\par

\section{Conclusions}
\label{sec:conclusion}
In this paper, we studied the problem of robust facial landmark detection regarding several aspects such as the use of datasets, evaluation metrics and methodology. 
Due to the performance saturation, we found that the widely used FR and NME measures can no longer effectively reflect the robustness of a model on several popular benchmarks. 
Therefore, we proposed several modifications to the current evaluation metrics and a novel method to make HRMs more robust. Our approach is based on the Wasserstein loss and involves training with smoother target heatmaps as well as a more precise coordinate sampling method using the barycenter of the output heatmaps.

\clearpage
%
%
\bibliographystyle{splncs04}
\bibliography{egbib}
\end{document}


\pagestyle{headings}
\mainmatter
\def\ECCVSubNumber{4031}  

\title{2D Wasserstein Loss for Robust Facial Landmark Detection \\ \vspace{5mm}
\large Supplementary Material} 

\titlerunning{ECCV-20 submission ID \ECCVSubNumber} 
\authorrunning{ECCV-20 submission ID \ECCVSubNumber} 
\author{Anonymous ECCV submission}
\institute{Paper ID \ECCVSubNumber}

\maketitle

\section{Additional Experimental Results}
In this section, we present additional experimental results that we are not able to demonstrate in the paper due to the space limit.

\subsection{Cross-dataset Validation} 
The landmark-wise CED of the protocol COFW$\rightarrow$AFLW is presented in Fig.~\ref{fig:cross_cofw}. Our method achieves a bigger improvement on COFW$\rightarrow$AFLW-All compared to COFW$\rightarrow$AFLW-Frontal. This is because AFLW-All contains non-frontal images, which is more challenging than AFLW-Frontal. 

On COFW$\rightarrow$AFLW-All (Fig.~\ref{fig:cross_cofw} (a)), by using Wasserstein loss, the NME performance can be improved by 0.13\% from 3.54\% to 3.41\%. Using CordConv can further improve the performance by 0.08\% to 3.33\%. Specifically, the improvement is significant at the NME from 6\% to 25\%. However, using big $\sigma$ will still decrease the local precision. We notice that the models using big $\sigma$ perform worse than the models using small $\sigma$ at NME=5\%.

\begin{figure}[]
\centering
\begin{subfigure}{.4\linewidth}
  \centering
  \includegraphics[width=\textwidth]{Fig/cofw-->aflwall.pdf}
  \caption{COFW$\rightarrow$AFLW-All}
\end{subfigure}
\begin{subfigure}{.4\linewidth}
  \centering
  \includegraphics[width=\textwidth]{Fig/cofw-->aflwfrontal.pdf}
  \caption{COFW$\rightarrow$AFLW-Frontal}
\end{subfigure}
\caption{Landmark-wise CED of COFW$\rightarrow$AFLW cross-validation with HRNet.}
\label{fig:cross_cofw}
\end{figure}

The landmark-wise CED of the protocol WFLW$\rightarrow$300VW is shown in Fig.~\ref{fig:cross_wflw}. We observe that by using Wasserstein Loss and CordConv, the HRNet trained on WFLW can be better generalized on the 300VW dataset. However, the improvement is less significant compared to the protocol 300W$\rightarrow$300VW. As discussed in the paper, the model trained on WFLW (with the dense annotation of 98 landmarks) has been already regularized by the strong landmark correlation among adjacent landmarks.

\begin{figure}[]
\centering
\begin{subfigure}{.32\textwidth}
  \centering
  \includegraphics[width=\linewidth]{Fig/wflw-->300vws3.pdf}
  \caption{WFLW$\rightarrow$300VW-S3}
\end{subfigure}
\begin{subfigure}{.32\textwidth}
  \centering
  \includegraphics[width=\linewidth]{Fig/wflw-->300vws2.pdf}
  \caption{WFLW$\rightarrow$300VW-S2}
\end{subfigure}
\begin{subfigure}{.32\textwidth}
  \centering
  \includegraphics[width=\linewidth]{Fig/wflw-->300vws1.pdf}
  \caption{WFLW$\rightarrow$300VW-S1}
\end{subfigure}
\caption{Cross-dataset validation of HRNet trained on WFLW (WFLW$\rightarrow$300VW). }
\label{fig:cross_wflw}
\end{figure}

\subsection{Comparison with State-of-the-art Using Traditional Evaluation Metrics}
We present comparison with state-of-the-art methods using traditional evaluation metrics on 300W (see Tab.~\ref{Tab:300w_valid_conventional}) and AFLW (see Tab.~\ref{Tab:aflw_valid_conventional}).
On both datasets, our model shows comparable performance to the state-of-the-art methods using traditional evaluation metrics. 
Here, using the Wasserstein loss only achieves a marginal improvement.
And using a larger $\sigma$ even slightly decreases the NME performance. 
As discussed in Sect.2 of the paper, the performance of vanilla HRNet has already reached a high level on these datasets. Thus, there are only very few challenging validation images for HRNet.
The NME is dominated by a large amount of small errors, which is the disadvantage of using a larger $\sigma$, and it 
can thus no longer reflect the robustness of the models. 
In the paper, the robustness of these models were validated by using cross-dataset validation.

\begin{table}[]
\centering
\begin{tabular}{llll}
\hline
Method & Common & Challenge & Full \\ \hline
PCD-CNN~\cite{kumar2018disentangling} & 3.67 & 7.62 & 4.44 \\
CPM+SBR~\cite{dong2018supervision} & 3.28 & 7.58 & 4.10 \\
SAN~\cite{dong2018san} & 3.34 & 6.60 & 3.98 \\
DAN~\cite{kowalski2017deep} & 3.19 & 5.24 & 3.59 \\
LAB~\cite{wayne2018lab} & 2.98 & \textbf{5.19} & 3.49 \\
DCFE~\cite{valle2018deeply} & \textbf{2.76} & 5.22 & \textbf{3.24} \\
DeCaFA~\cite{Dapogny_2019_decafa} & 2.93 & 5.26 & 3.39 \\ \hline
HRNet, $\sigma=1$, $L2$ & 2.91 & 5.11 & 3.34 \\
$\sigma=3$, $L2$ & 3.05 & 5.28 & 3.49 \\
$\sigma=1$, W Loss & 2.85 & 5.13 & 3.29 \\
$\sigma=3$, W Loss & 3.01 & 5.30 & 3.46 \\
$\sigma=1$, W Loss, CC & \textbf{2.81} & \textbf{5.08} & \textbf{3.26} \\
$\sigma=3$, W Loss, CC & 2.95 & 5.22 & 3.39 \\ \hline
\end{tabular}
\caption{NME (\%) comparison on 300W. W Loss - Wasserstein Loss. CC - CoordConv.}
\label{Tab:300w_valid_conventional}
\end{table}

\begin{table}[]
\centering
\begin{tabular}{c|c|c|c|c|c|c}
\hline
Method & SAN~\cite{dong2018san} & DSRN~\cite{miao2018direct} & Wing-Loss~\cite{feng2018wing} & SA~\cite{liu2019semantic} & LAB w/o B~\cite{wayne2018lab} & ODN~\cite{zhu2019robust} \\ \hline
Frontal & 1.85 & - & - & - & 1.62 & \textbf{1.38} \\
All & 1.91 & 1.86 & \textbf{1.47} & 1.60 & 1.85 & 1.63 \\ \hline \hline
HRNet & $\sigma=1.5$, $L2$ & $\sigma=3$, $L2$ & $\sigma=1.5$, W & $\sigma=3$, W & $\sigma=1.5$, W, CC  & $\sigma=3$, W, CC \\ \hline
Frontal & 1.46 & 1.44 & 1.39 & 1.43 & \textbf{1.37} & 1.43 \\
All & 1.57 & 1.57 & \textbf{1.51} & 1.58 & \textbf{1.51} & 1.57 \\ \hline
\end{tabular}
\caption{NME(\%) performance comparision on AFLW. W - Wasserstein Loss. CC - CoordConv. B - Boundary}
\label{Tab:aflw_valid_conventional}
\end{table}

\clearpage
%
%
\bibliographystyle{splncs04}
\bibliography{egbib}